# Simplified Swarm Optimization for Bi-Objection Active Reliability Redundancy Allocation Problems

Wei-Chang Yeh
Integration and Collaboration Laboratory
Department of Industrial Engineering and Engineering Management
National Tsing Hua University
yeh@ieee.org

**Abstract**: The reliability redundancy allocation problem (RRAP) is a well-known tool in system design, development, and management. The RRAP is always modeled as a nonlinear mixed-integer non-deterministic polynomial-time hardness (NP-hard) problem. To maximize the system reliability, the integer (component active redundancy level) and real variables (component reliability) must be determined to ensure that the cost limit and some nonlinear constraints are satisfied. In this study, a bi-objective RRAP is formulated by changing the cost constraint as a new goal, because it is necessary to balance the reliability and cost impact for the entire system in practical applications. To solve the proposed problem, a new simplified swarm optimization (SSO) with a penalty function, a real one-type solution structure, a number-based self-adaptive new update mechanism, a constrained nondominated-solution selection, and a new *pBest* replacement policy is developed in terms of these structures selected from full-factorial design to find the Pareto solutions efficiently and effectively. The proposed SSO outperforms several metaheuristic state-of-the-art algorithms, e.g., nondominated sorting genetic algorithm II (NSGA-II) and multi-objective particle swarm optimization (MOPSO), according to experimental results for four benchmark problems involving the bi-objective active RRAP.

## 1. INTRODUCTION

Network reliability is the probability that a network operates successfully under predefined and required conditions. The ever-advancing nature of technology has led to the curation of networks, which have a pervasive influence on the modern world and is a critical element in its makeshift. Thus, network reliability is arguably more relevant than before. Network reliability has been applied to evaluate the performance of various types of systems and networks such as the internet of things [1], data mining techniques [2], traffic networks [3], and cloud computing [4] in designing an optimizing and reliable system architecture.

Reliability design essentially aims to enhance system performance under a limited budget, and has become a popular research topic in the study of network reliability. There are two key different reliability design techniques that have been proposed: to increase the component reliability directly or to use redundant components simultaneously. The latter is more economical and practical if the cost, weight, and volume of components are taken into consideration and is a considerably more popular research topic. Hence, more and more researchers study the latter from many different perspectives than that of the former.

All reliability design problems that employ the technique of using redundant components simultaneously can be formulated in a general nonlinear mixed-integer programming model [5-19] and its details are discussed in Section 2.3. In the literature, there are three kinds of reliability design problems: redundancy allocation problem (RAP) in which reliability variables are known and redundancy variables are unknown, reliability allocation problem in which redundancy variables are known and reliability variables are unknown, as well as the reliability–redundancy allocation problem (RRAP) in which both reliability variables and redundancy variables are unknown. Compared to the other two problems, RRAP is more reasonable and more difficult to solve. Thus, the RRAP is considered in this study.

Because many distinct types of systems are employed in various fields, there are different categories of RRAPs with different redundancy strategies, i.e., active strategies [5, 6, 8, 9, 11, 13, 14, 17-19], standby strategies [7, 10, 15], and mixed strategies [12, 16]. The differences between these strategies are discussed briefly below.

(1) Active RRAP [5, 6, 8, 9, 11, 13, 14, 17-19]: all components are full operated, but only one is required for system operation.
(2) Standby redundancy strategies:
   a. Cold-standby RRAP [7, 10, 15]: all redundancy components are unpowered, and their

failure rate is always assumed to be zero when they are in standby.

   b. Hot-standby RRAP: all redundancy components are full operated, and their mathematical formulation is identical to that for the active strategy.

   c. Warm-standby RRAP: all redundancy components are partially operated.

(3) Mixed RRAP [12, 16]: the redundancy components can be of any of the types listed above.

According to the characteristics of the components used in RRAPs, there are various RRAPs, e.g., a fuzzy RRAP to overcome the uncertainty of component parameters [11], a heterogeneous RRAP to use different types of components in the subsystem [12], and a multi-state $k$-out-of-$n$ RRAP with reparable components [19]. Regardless of the standby strategy, the redundancy components must replace the failed main component to prevent the entire system from crashing. The active RRAP with the active redundancy strategy is the most widely used method among all the aforementioned redundancy strategies. Hence, it was adopted in the present study.

To solve this well-known NP-hard problem, various artificial-intelligence techniques have been widely studied in the past decades for numerous systems with diverse considerations [5-19]. Examples of such techniques include stochastic fractal search (SFS) [17], genetic algorithm (GA) [10], a hybrid algorithm of GA and PSO [13], particle swarm optimization (PSO) [9], stochastic perturbation PSO [12], the artificial bee colony algorithm [6], cuckoo search [18], particle-based simplified swarm optimization (SSO) [5], and SSO with the boundary-search [7, 8]. Among these algorithms, it appears that SSO with the boundary-search proposed in [7, 8] are the best for cold-standby RRAP and active RRAP, respectively.

With the successful development of multi-objective artificial-intelligence algorithms, there has been growing research interest in solving the RRAP as a bi-objective optimization problem in recent years. Ardakan and Hamadani [10] implemented nondominated sorting genetic algorithm II (NSGA-II) to solve bi-objective cold-standby RRAPs. Garg et al. investigated the bi-objective fuzzy RRAP in a series-parallel system by converting it to a single-objective RRAP [9]. Raouf and Pourtakdoust [13] solved a bi-objective RRAP by considering the type of component as a variable, with the reliability and the cost as objectives. Muhuri et al. [14] adapted the elitist NSGA-II to solve the bi-objective type-2 fuzzy RRAP.

Ardakan and Rezvan [15] proposed NSGA-II to solve a bi-objective cold-standby RRAP. Wang et al. [16] solved the mixed RRAP for a multi-type production system via NSGA-II.

This paper presents a bi-objective active RRAP to maximize the reliability and minimize the cost. As indicated by the foregoing discussion, NSGA-II is one of the most widely used state-of-the-art multi-objective algorithms for the bi-objective RRAP. To explore the bi-objective active RRAP to have more multi-objective algorithms rather than depending on NSGA-II, the SSO proposed by Yeh [7, 8] is extended. The resulting algorithm, which is called multi-objective SSO (MOSSO), is used to solve the bi-objective active RRAP. The performance of the proposed MOSSO is confirmed by comparing the experimental results with those for NSGA-II and multi-objective PSO (MOPSO), which is also a state-of-the-art multi-objective algorithm, for four well-known RRAP benchmark problems.

The remainder of this paper is organized as follows. Section 2 provides the required background for the proposed MOSSO, including the SSO, crowding distance, and repository, as well as the formulations of the four benchmark RRAPs, Pareto front, and nondominated solutions. Section 3 presents the proposed MOSSO for the bi-objective active RRAP. Section 4 presents a complete comparative experiment in which the four RRAP benchmark problems were solved using two state-of-art algorithms: NSGA-II and MOPSO. Section 5 presents discussions and conclusions.

## 2. BACKGROUND OF SSO, RRAP, AND MULTI-OBJECTIVE PROBLEMS

The SSO is the basis of the proposed MOSSO for solving the proposed bi-objective active RRAP. The crowding distance used in NSGA-II and the repository implemented in MOPSO both are adapted in the proposed MOSSO. Before the proposed MOSSO is described, a general background regarding the RRAP, multi-objective problems, the Pareto front, SSO, the crowding distance, NSGA-II, the repository, and MOPSO, are presented.

### 2.1 The RRAP and its four benchmarks

A nonlinear mixed-integer programming problem for RRAP and four RRAP benchmark problems can be modeled generally as follows [5-19].

$$\text{Maximize} \quad R_s(\mathbf{n}, \mathbf{r}) \tag{1}$$

$$\text{Subject to} \quad g_v(\mathbf{n}, \mathbf{r}) \leq V_{ub} \tag{2}$$

$$g_c(\mathbf{n}, \mathbf{r}) \leq C_{ub} \tag{3}$$

$$g_w(\mathbf{n}, \mathbf{r}) \leq W_{ub} \tag{4}$$

$$\mathbf{n}_{lb} \leq \mathbf{n} = (n_1, n_2, \ldots, n_{N_{var}}) \leq \mathbf{n}_{ub} \tag{5}$$

$$\mathbf{r}_{lb} \leq \mathbf{r} = (r_1, r_2, \ldots, r_{N_{var}}) \leq \mathbf{r}_{ub}. \tag{6}$$

where

$N_{var}$ : number of discrete/continuous variables (i.e., subsystems), i.e., $N_{var} = N_{sub}$. Note that the reliability of each component is identical in each subsystem.

$n_i$    the $i$th redundancy variable for $i = 1, 2, \ldots, N_{var}$.

$r_i$    the $i$th reliability variable for $i = 1, 2, \ldots, N_{var}$.

$\mathbf{n}$    $\mathbf{n} = (n_1, n_2, \ldots, n_{N_{var}})$.

$\mathbf{r}$    $\mathbf{r} = (r_1, r_2, \ldots, r_{N_{var}})$.

$R_s(\mathbf{n}, \mathbf{r})$    The RRAP reliability under $\mathbf{n}$ and $\mathbf{r}$.

$g_v(\mathbf{n}, \mathbf{r})$    The constraint of the volume under $\mathbf{n}$ and $\mathbf{r}$ in the RRAP.

$g_c(\mathbf{n}, \mathbf{r})$    The constraint of the cost under $\mathbf{n}$ and $\mathbf{r}$ in the RRAP.

$g_w(\mathbf{n}, \mathbf{r})$    The constraint of the weight under $\mathbf{n}$ and $\mathbf{r}$ in the RRAP.

$V_{ub}, C_{ub}, W_{ub}$    The upper bound of the constraints related to the volume, cost, and weight, respectively, in the RRAP.

The reliability functions, number of subsystems, network structures of four benchmark RRAPs are provided in Table 1 together with the corresponding parameters listed in Table 2.

**Table 1.** The reliability functions and system structures of four RRAP benchmark problems.

| ID | $N_{sub}$ | $R_s(\mathbf{n}, \mathbf{r})$ | Structure |
|---|---|---|---|
| 1 | 5 | $\prod_{i=1}^{N_{sub}} [1 - (1 - r_i)^{n_i}]$ | 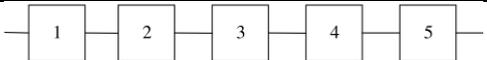 the series system |

| 2 | 5 | $1-(1-r_1^{n_1}r_2^{n_2})\{1-[1-(1-r_3^{n_3})(1-r_4^{n_4})]r_5^{n_5}\}$ | 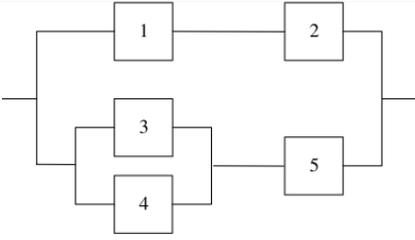 the series-parallel system |
|---|---|---|---|
| 3 | 5 | $r_1^{n_1} \cdot r_2^{n_2} + r_3^{n_3} \cdot r_4^{n_4} + r_1^{n_1} \cdot r_4^{n_4} \cdot r_5^{n_5} + r_2^{n_2} \cdot r_3^{n_3} \cdot r_5^{n_5} -$ $r_1^{n_1} \cdot r_2^{n_2} \cdot r_3^{n_3} \cdot r_4^{n_4} - r_1^{n_1} \cdot r_2^{n_2} \cdot r_3^{n_3} \cdot r_5^{n_5} - r_1^{n_1} \cdot r_2^{n_2} \cdot r_4^{n_4} \cdot r_5^{n_5} -$ $r_1^{n_1} \cdot r_3^{n_3} \cdot r_4^{n_4} \cdot r_5^{n_5} - r_2^{n_2} \cdot r_3^{n_3} \cdot r_4^{n_4} \cdot r_5^{n_5} + 2 r_1^{n_1} \cdot r_2^{n_2} \cdot r_3^{n_3} \cdot r_4^{n_4} \cdot r_5^{n_5}$ | 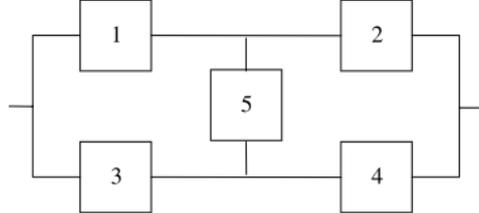 the bridge system |
| 4 | 4 | $\prod_{i=1}^{N_{sub}}[1-(1-r_i)^{n_i}]$ | 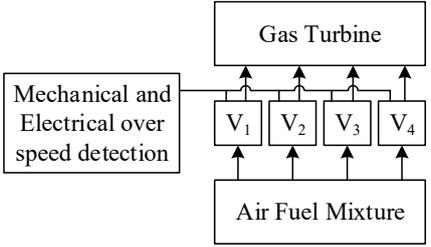 the overspeed protection of a gas turbine system |

**Table 2.** The information of all parameters of four RRAP benchmark problems.

| ID | Subsystem $i$ | $\alpha_i \cdot 10^5$ | $\beta_i$ | $w_i v_i^2$ | $w_i$ | $V_{ub}$ | $C_{ub}$ | $W_{ub}$ |
|---|---|---|---|---|---|---|---|---|
| 1, 3 | 1 | 2.330 | 1.5 | 1 | 7 | | | |
| | 2 | 1.450 | 1.5 | 2 | 8 | | | |
| | 3 | 0.541 | 1.5 | 3 | 8 | 110 | 175 | 200 |
| | 4 | 8.050 | 1.5 | 4 | 6 | | | |
| | 5 | 1.950 | 1.5 | 2 | 9 | | | |
| 2 | 1 | 2.500 | 1.5 | 2 | 3.5 | | | |
| | 2 | 1.450 | 1.5 | 4 | 4.0 | | | |
| | 3 | 0.541 | 1.5 | 5 | 4.0 | 180 | 175 | 100 |
| | 4 | 0.541 | 1.5 | 8 | 3.5 | | | |
| | 5 | 2.100 | 1.5 | 4 | 4.5 | | | |
| 4 | 1 | 1 | 1.5 | 1 | 6 | | | |
| | 2 | 2.3 | 1.5 | 2 | 6 | 250.0 | 400.0 | 500.0 |
| | 3 | 0.3 | 1.5 | 3 | 8 | | | |
| | 4 | 2.3 | 1.5 | 2 | 7 | | | |

There is only one objection of the traditional RRAP, i.e., Eq. (5), and it is to maximize $R_s(\mathbf{n}, \mathbf{r})$ by having $\mathbf{n}=(n_1, n_2, \ldots, n_{Nvar})$ and $\mathbf{r}=(r_1, r_2, \ldots, r_{Nvar})$.

In the active RRAP, all components are assumed non-repairable, binary-states (operational or failed), identical in the same subsystem, exponential distribution of the time-to-failure, and with predetermined reliability. Also, the sates of different components are *s*-independent, i.e., the occurrence of one component failure does not affect the occurrence of the other failure event.

**2.2 Pareto solutions and Pareto front**

Modern applications are increasingly focused on multiple objectives, e.g., internet of things [20], cloud computing [21], and sensor networks [22]; thus, multi-objective problems are more important than ever before.

Solution $X_i$ is a dominated solution $X_j$ if the fitness values of $X_i$ is better than that of $X_j$. A global (feasible) nondominated solution is also called a Pareto solution. A local (feasible) nondominated solution is a solution that currently dominates the others and may be dominated after generations.

There are numerous global nondominated solutions if there is no solution such that all the objective values are the best. In the following example, two objective functions need to be maximized, where $f_i(X)$ represents the $i$th objective value of $X$ in the multi-objective problem:

$$\text{Maximize} \quad f_1(x_1, x_2) = x_1 + x_2 \tag{7}$$

$$\text{Maximize} \quad f_2(x_1, x_2) = x_1 - x_2 \tag{8}$$

$$\text{s.t.} \quad 0 \leq x_1, x_2 \leq 1. \tag{9}$$

The maximum objective-function values of $f_1(x_1, x_2)$ and $f_2(x_1, x_2)$ are 2 from $X_1 = (x_1, x_2) = (1, 1)$ and 1 from $X_2 = (x_1, x_2) = (1, 0)$, respectively. Hence, there is no solution such that both objective-function values are maximized, and there are many nondominated solutions, e.g., $f_1(X_1) = 2 \geq f_1(X_2) = 1$ and $f_2(X_1) = 0 \leq f_2(X_2) = 1$.

There are different methods for solving multi-objective problems, and they can be classified into two: 1) transformation methods, which transform the multi-objective problem into a single-objective problem, and 2) Pareto methods, which attempt to obtain the Pareto front formed as a curve by global nondominated solutions.

The transformation methods are simpler than the Pareto methods and use techniques to combine all the objective functions, e.g., the weighted method [23] and the fuzzy set [24], for transforming the multi-objective problem into a single-objective problem, which can be solved using any single-objective method. The solutions obtained via the transformation methods are not always local nondominated solutions, and their function values are always the same. As mentioned previously, if there is no solution that can optimize all the objective functions, the nondominated solution is not the only one.

Thus, most nondominated solutions cannot be found via transformation methods, and some nondominated solutions may be better than the optimal solutions obtained via transformation methods. Transformation methods are not suitable if more than one solution is needed to make decisions. Thus, the remainder of this study is focused only on the Pareto methods.

The Pareto methods involve finding the Pareto front, which does not need to be continuous. However, rather than one optimum, the number of solutions in a curve maybe infinite, and it is impossible to find all the global nondominated solutions. Thus, we attempt to find sufficient nondominated solutions to cover the curve uniformly. Moreover, similar to the single-objective problem, it very difficult to find a good solution that is equal to or very close to the optimum at the beginning of all artificial intelligence (AI) algorithms. Hence, all the multi-objective AI algorithms attempt to approach and uniformly cover the Pareto front to the greatest extent possible during a series of update processes.

**2.3 Overview of SSO**

First proposed by Yeh [25] in 2008, the simplified swarm optimization (SSO) algorithm is, in essence, a combination of swarm intelligence and evolutionary computation. Numerous studies have shown the competency of SSO and the critical role the algorithm plays in a myriad of research areas including: the RAP which was solved by the first orthogonal SSO [25], the RRAP using the SSO hybrid with the PSO [5], the disassembly sequencing problem in the green supply chain domain with the first self-adaptive SSO [26, 27], the training of artificial neural network in mining time series data which is based on the first continuous SSO [28], the high-dimensional numerical continuous functions that use an improved continuous SSO in [29], SSO with the macroscopic indeterminacy in [30], SSO combined with the glowworm swarm optimization in [31], the RFID network problem in health care management [32], the dispatch problems solving by hybrid bacterial foraging and SSO [33] and by gradient-based SSO [34], and the network security problem in detecting network intrusions [35], etc. Experimental results have confirmed that the SSO and its variants outperform PSO, GA, EDA, and ANN in RAP problems [25, 36, 37], RRAP problems [5, 7, 8], and other problems [26-35, 37-42].

Let $N_{sol}$ be the number of solutions, $X_i = (x_{i,1}, x_{i,2}, \ldots, x_{i,Nvar})$ be the solution $i$, $P_i = (p_{i,1}, p_{i,2}, \ldots, p_{i,Nvar})$ be the best solution $i$ in the its own evolution history and it is also called the *pBest* of the solution $i$, and $G = (g_1, g_2, \ldots, g_{Nvar})$ be the best solution found so far (i.e., *gBest*), where $x_{i,j}$, $p_{i,j}$, and $g_j$ are the *j*th variables in the $X_i$, $P_i$, and $G$, respectively, for $i = 1, 2, \ldots, N_{sol}$ and $j = 1, 2, \ldots, N_{var}$. Break even if there is a tie in the best solutions or the best solutions are updated from the *i*th solution.

Similar to all major AI, in the first generation of SSO, all solutions are randomly initialized [5, 7, 8, 25-43]. The basic idea of the update mechanism SSO in updating $x_{i,j}$ is based on the following stepwise functions [5, 7, 8, 25-43].

$$x_{i,j} = \begin{cases} g_j & \text{if } \rho_{[0,1]} \in [0, C_g) \\ p_{i,j} & \text{if } \rho_{[0,1]} \in [C_g, C_p) \\ x_{i,j} & \text{if } \rho_{[0,1]} \in [C_p, C_w) \\ x & \text{otherwise} \end{cases}, \quad (10)$$

where the value $\rho_{[0,1]} \in [0, 1]$ is generated randomly. In the first item of Eq. (10), the new $x_{i,j}$ is taken from $g_j$ with probabilities $C_g$, from $p_{i,j}$ with probability $C_p - C_g$, from itself with probability $C_w - C_p$, and from a random generated feasible value $x$ with probability $1 - C_w$.

The $P_i$ and $G$ are updated accordingly based on Eqs. (11) and (12) if we want to maximize our objective function, i.e., survival-of-the-fittest, when the more optimal values of $P_i$ and $G$ are obtained:

$$P_i = \begin{cases} X_i & \text{if } F(P_i) < F(X_i) \\ P_i & \text{otherwise} \end{cases}, \quad (11)$$

$$G = \begin{cases} X_i & \text{if } F(G) < F(X_i) \\ G & \text{otherwise} \end{cases}. \quad (12)$$

The update mechanism is the core of all AI algorithms and each of them has its own unique update mechanism. Compared to the update mechanisms of other AI algorithms, e.g., PSO, ABC, differential evolution (DE), ant colony optimization, cuckoo search, Eq. (10) is much simpler. Hence, SSO is easy to be understood, coded, and implemented.

The flexible nature of the stepwise function listed in Eq. (10) highlights the malleability of SSO and underscores how it allows for ease of modification and customization with various practical applications, as discussed in the beginning of this subsection. Therefore, varying customized SSO can be proposed for diverse problems from the no free lunch theorem. For example, the continuous SSO which mainly focuses on problems with real variables [28, 29], the one-variable SSO which only updates one variable for each solution rather than updating all variables like the traditional SSO [28, 39], the 3-item SSO which removes the role of *pBest* in Eq. (10) for some problems [29, 41, 42], elite SSO which selects the best $N_{var}$ solutions among the parent generations and the offspring to the next generation [7, 38, 40]. However, the stepwise function will always remain the basis of SSO update mechanism [5, 7, 8, 25-43].

Let $N_{gen}$ be the number of generations. The pseudocode of SSO is listed below.

**SSO PROCEDURE** [25, 29, 43]

**STEP S0.** Generate solutions $P_1 = X_1$, $P_2 = X_2$, …, $P_{Nsol} = X_{Nsol}$ randomly, calculate their fitness values, let $t = i = 1$, and find $G$ such that $G$ has the best fitness among all solutions.

**STEP S1.** Update $X_i$ based on Eq. (10).

**STEP S2.** Let $P_i = X_i$ if the fitness of $X_i$ is better than that of $P_i$. Otherwise, proceed to STEP S4.

**STEP S3.** Let $G = X_i$ if the fitness of $X_i$ is better than that of $G$.

**STEP S4.** Let $i = i + 1$ and proceed to STEP S1 if $i < N_{sol}$.

**STEP S5.** Let $t = t + 1$, $i = 1$, and proceed to STEP S1 if $t < N_{gen}$. Otherwise, halt and $G$ is the solution we need.

From the above, SSO is simple and efficient. Hence, it is revised in this study for the bi-objective active RRAP.

## 2.4. MOPSO and repository

MOPSO is proposed by Coello and Lechuga in [44] based on PSO which is swarm intelligence algorithm proposed by Kennedy and Eberhart [45, 46]. There are three key characteristics of MOPSO that PSO does not have, these include:

1) MOPSO has a special limited space called the repository to store found local nondominated solutions, which was first used in the adaptive grid of Pareto archived evolution strategy [44].

2) The *gBest* in PSO is replaced with a local nondominated solution selected randomly from the repository.

3) The mutations used in the GA are adapted in MOPSO to solve the premature problems in occurred in MOPSO.

The above first two concepts are also tested in the proposed MOSSO and their integration methods are discussed in Section 3.

Let $N_{rep}$ be the maximum number of local nondominated solutions stored in the repository. To maintain a constant $N_{rep}$, the adaptive grid procedure must be implemented in MOPSO in a manner that solutions located in more colonized regions of objective space are given priority over those lying in less colonized areas. Moreover, for one of the updated solutions, say $X_i$, and its *pBest* $P_i$ will be selected randomly to be the new *pBest* if $X_i$ and $P_i$ are not dominated by each other.

The main process of the MOPSO is described below.

**STEP P0.** Randomly generate all the $N_{sol}$ solutions (called "particles" or "positions" in PSO and MOPSO): $X_1, X_2, \ldots, X_{N_{sol}}$ and velocity $V_1, V_2, \ldots, V_{N_{sol}}$. Let the repository $\Omega$ be the set of all local nondominated solutions among $X_1, X_2, \ldots, X_{N_{sol}}$.

**STEP P1.** Update $V_i$ and $X_i$ according to PSO, except that the *gBest* is selected randomly from $\Omega$.

**STEP P2.** If $X_i$ is dominated by $P_i$, proceed to STEP P7.

**STEP P3.** If $X_i$ is dominated by $P_i$, let $X_i = P_i$ and proceed to STEP P6.

**STEP P4.** If $X_i$ and $P_i$ are not dominated by each other, select one solution from $X_i$ and $P_i$ randomly to be the new $P_i$.

**STEP P5.** If $X_i$ is the new $P_i$, proceed to STEP P6. Otherwise, proceed to STEP P7.

**STEP P6.** Let Ω = Ω ∪ {$X_i$} and then remove these dominated solutions from Ω. If |Ω| > $N_{sol}$, the adaptive grid procedure is implemented to keep |Ω| equal to $N_{sol}$ by discarding extra solutions.

**STEP P7.** Let $i = i + 1$ and proceed to STEP P1 if $i < N_{sol}$.

**STEP P8.** Let $t = t + 1$, $i = 1$, and proceed to STEP P1 if $t < N_{gen}$. Otherwise, halt and all solutions in Ω are local nondominated solutions.

**2.5 NSGA-II and crowding distance**

As discussed in the previous subsection, we need the local nondominated solutions to cover the Pareto front uniformly, e.g., they cannot gather in some part of the Pareto front. To force the updated solutions to be as diverse as possible, the crowding distance proposed for NSGA-II by K. Deb in 2002 is adapted in the proposed MOSSO. NSGA-II is a well-known algorithm for finding multiple Pareto solutions according to the crowding distance and the nondominated sorting in multi-objective optimization problems.

The crowding distance is a diversity-preserving mechanism and is adopted in this study to maintain a constant number of local nondominated solutions in the repository. It determines the impact of the solution diversity of the related neighborhood of the Pareto front such that a longer crowding distance makes the number of local nondominated solutions more reasonable to keep for the next generation.

Let $f_{i,k-1}$ and $f_{i,k+1}$ be the fitness values of two solutions such that $f_{i,k+1}$ is the smallest fitness value larger than $f_i(X_k)$ and $f_{i,k-1}$ is the largest fitness value smaller than $f_i(X_k)$ for $i = 1, 2, \ldots, N_{obj}$, where $N_{obj}$ represents the number of objective functions. The crowding distance of each solution, e.g., $X_k$, is an estimation of the solution density along the Pareto front. It is calculated as the summation of the individual normalized Euclidean distances of the two neighboring solutions, i.e., $f_{i,k-1}$ and $f_{i,k+1}$ for the $k$th objective function, corresponding to each objective based on the following equation:

$$D(X_i) = \sqrt{\sum_{i=1}^{N_{obj}}\left[\left(\frac{f_i(X_k) - f_{i,k-1}}{f_{i,max} - f_{i,min}}\right)^2 + \left(\frac{f_{i,k+1} - f_i(X_k)}{f_{i,max} - f_{i,min}}\right)^2\right]}, \qquad (13)$$

where $f_{i,\min}$ and $f_{i,\max}$ represent the minimum and maximum values of the $i$th objective function.

NSGA-II is based on the traditional GA, including the implementation of the crossover operator and mutation operator to update solutions (chromosomes) [10, 14-16]. Let $X_{t,1}, X_{t,2}, \ldots, X_{t,\text{Nsol}}$ be solutions and $X_{t,\text{Nsol}+1}, X_{t,\text{Nsol}+2}$, be offspring (solutions) generated by parents randomly selected from $\{X_{t,1}, X_{t,2}, \ldots, X_{t,\text{Nsol}}\}$ in the generation $t$. The major difference between the GA and NSGA-II is the method of selecting solutions $X_{t+1,1}, X_{t+1,2}, \ldots, X_{t+1,\text{Nsol}}$ from $\Omega = \{X_{t,1}, X_{t,2}, \ldots, X_{t,2\text{Nsol}}\}$ for generation $(t+1)$. Hence, there are always $N_{\text{sol}}$, $2 \times N_{\text{sol}}$, and $N_{\text{sol}}$ solutions before the update, after the update, and after the selection process, respectively.

In NSGA-II, all nondominated solutions are selected first via nondominated sorting. Let the total number of local nondominated solutions be $N_{\text{lns}}$. If $N_{\text{lns}} > N_{\text{sol}}$, to maintain a constant number of solutions, the $(N_{\text{lns}} - N_{\text{sol}})$ nondominated solutions with the shortest crowding distances are discarded, and the remaining $N_{\text{sol}}$ nondominated solutions are kept for generation $(t+1)$.

Let $\Omega = \{X_{t,1}, X_{t,2}, \ldots, X_{t,2\text{Nsol}}\}$ be the set of all parents and offspring, $\Omega_i$ be the set of local nondominated solutions in $\Omega - \bigcup_{k=1}^{i-1}\Omega_k$, and $j$ be the smallest index such that $\sum_{k=1}^{j}|\Omega_k| \geq N_{\text{sol}} = N_{\text{lns}}$. The foregoing process for finding all the solutions with different rankings is called "nondominated sorting" in NSGA-II. If $N_{\text{sol}} < \sum_{k=1}^{j}|\Omega_k|$, the crowding distance must be calculated for all solutions in $\Omega_j$ using Eq. (13), and $(\sum_{k=1}^{j}|\Omega_k| - N_{\text{sol}})$ solutions are discarded from $\Omega_j$.

The main steps of NSGA-II are as follows.

**STEP G0.** Generate all $N_{\text{sol}}$ solutions (called chromosomes in GA or NSGE-II) randomly. Let $\Omega_{\text{old}} = \{X_{t,1}, X_{t,2}, \ldots, X_{t,\text{Nsol}}\}$ and $t = 1$.

**STEP G1.** Apply the crossover operator and the mutation operator to update the solution selected randomly from $\Omega_{\text{old}}$, and let $\Omega_{\text{new}} = \{X_{t,\text{Nsol}+1}, X_{t,\text{Nsol}+2}, \ldots, X_{t,2\text{Nsol}}\}$ be these new offspring. Additionally, let $\Omega = \Omega_{\text{old}} \cup \Omega_{\text{new}} = \{X_{t,1}, X_{t,2}, \ldots, X_{t,2\text{Nsol}}\}$.

**STEP G2.** Use the nondominated sorting to rank the solutions in $\Omega$ to find $\Omega_i = \Omega - \bigcup_{k=1}^{i-1} \Omega_k$. Let $j$ be the smallest index such that $\sum_{k=1}^{j} |\Omega_k| \geq N_{sol} = N_{lns}$.

**STEP G3.** Calculate the crowding distance for each solution in $\Omega_j$. Let $\Omega_{del} \subseteq \Omega_j$ be the set of $\sum_{k=1}^{j} |\Omega_k| - N_{sol}$ solutions with the shortest crowding distances, and let $\Omega_{old} = \bigcup_{k=1}^{j} \Omega_k - \Omega_{del}$.

**STEP G4.** If $t < N_{gen}$, let $t = t + 1$ and proceed to STEP G1. Otherwise, halt and all nondominated solutions in $\Omega_{old}$ form the local Pareto front.

## 3. NEW MOSSO FOR BI-OBJECTIVE ACTIVE RRAP

The major parts of the proposed MOSSO are introduced in this section, including the one-type solution structure to combine the redundancy variable and the reliability variable, the two penalty functions for the two objective functions, the novel number-based self-adaptive update mechanism employing the repository and the constrained nondominated-solution selection to select the local nondominated solutions in the early generations, the crowding distance to maintain a constant number of local nondominated solutions in the repository, the new *pBest* replacement policy to determine whether *pBest* needs to be replaced, and the full-factorial design to help to determine the best structure and components.

### 3.1 The one-type solution structure

All RRAPs include both the integer redundancy variables and the real reliability variables. Hence, it is very common to use two different update mechanisms to update these two different types of variables.

To reduce the run time and improve the solution quality, Yeh first proposed the solution structure by combining two different types of variables into a real variable in the cold-standby RRAP [7]. In such real variable, the integer denotes the number of redundancies and the value after the decimal point is the $\lambda$ (a parameter in the cold-standby RRAP) of the subsystem [7]. For example, solution $X = (x_1, x_2, x_3, x_4) =$

(3.01999, 3.00587, 2.00836, 1.0368) denotes that $n_1 = n_2 = 3$, $n_3 = 2$, $n_4 = 1$, $\lambda_1 = 0.01999$, $\lambda_2 = 0.00587$, $\lambda_3 = 0.00836$, and $\lambda_4 = 0.0368$.

Here, the one-type solution structure used in the cold-standby RRAP is revised in the proposed bi-objective active RRAP. In such structure, the integer part is still the number of redundancies but the digits after the decimal point is the reliability of the component corresponding to the integer part in the bi-objective active RRAP. For example, redundancy variable **n** = (3, 3, 2, 1) and reliability variable **r** = (0.91999, 0.90587, 0.90836, 0.9368) is combined to be the solution $X$ = (3.91999, 3.90587, 2.90836, 1.9368).

## 3.2 The penalty function

To improve infeasible solutions near the border in the feasible solution space, there is always a penalty function implemented if any constraint among Eqs. (2)–(4) is not satisfied. As in lately published papers [5, 6, 36, 47], the following penalty function $F_R(X)$ is implemented to replace $R_s(X)$ in RRAP, where $X = (\mathbf{n}, \mathbf{r})$:

$$F_R(X) = R_s(X) \times (\text{Min}\{R_s(X)/R_{lb}, V_{ub}/g_v(X), W_{ub}/g_w(X), C_{ub}/g_c(X)\})^3. \tag{14}$$

To meet the characteristic of the proposed bi-objective active RRAP, a new additional penalty function $F_C(X)$ is proposed to replace $g_c(X)$:

$$F_C(X) = g_c(X) / (\text{Min}\{R_s(X)/R_{lb}, V_{ub}/g_v(X), W_{ub}/g_w(X), C_{ub}/g_c(X)\})^3. \tag{15}$$

## 3.3 Number-based self-adaptive update mechanism

The traditional update mechanism in SSO is significantly revised for the bi-objective model in the proposed MOSSO. Let local nondominated solution $G = (g_1, g_2, \ldots, g_{N_{var}})$ be selected randomly from the repository, which is an extra set to store a limited number of nondominated solutions. In the proposed MOSSO, each variable in the solution $X_i = (x_{i,1}, x_{i,2}, \ldots, x_{i,N_{var}})$ is updated according to the following stepwise function:

$$x_{i,j} = \begin{cases} g_j & \text{if } \rho_{[0,1]} \in [0., 0.8\sqrt[3]{\frac{N_{\text{Ins},t}}{N_{\text{gns}}}}) \\ p_{i,j} & \text{if } \rho_{[0,1]} \in [0.8\sqrt[3]{\frac{N_{\text{Ins},t}}{N_{\text{gns}}}}, C_p) \\ x_{i,j} & \text{if } \rho_{[0,1]} \in [C_p, C_w) \\ x & \text{if } \rho_{[0,1]} \in [C_w, 1] \end{cases}, \quad (16)$$

where $p_{i,j}$ represents the $j$th variable of the *pBest* of $X_i$, the random variable $\rho_{[0,1]}$ is generated within the interval [0, 1] uniformly, the random variable $x$ is generated within the interval $[L_j, U_j]$ uniformly, $i$ = 1, 2, …, $N_{\text{sol}}$ is the solution index, and $j$ = 1, 2, …, $N_{\text{var}}$ is the variable index.

There are two differences between the proposed update mechanism and that of the traditional SSO.

1) In the proposed update mechanism, the best solution used is selected from the repository randomly, not deterministically. This is because the local nondominated solutions are not dominated by each other in the repository, i.e., *gBest* is not only one. In single-objective problems, the fitness of the *gBest* in the current generation is always the same; hence, the definition of the best solution is revised here.

2) In the proposed update mechanism, the parameter $C_g = 0.8\sqrt[3]{\frac{N_{\text{Ins},t}}{N_{\text{gns}}}}$ is a self-adaptive number (whereas it is a fixed number in the traditional SSO). $C_g$ increases if the number of nondominated numbers increases. This is mainly because in the early stage, the number of nondominated numbers is small, and the solution diversity is insufficient. Hence, either no new nondominated solutions are found, or the new nondominated solutions generated according to Eq. (16) are too close to *G*. In the proposed MOSSO, for preventing these two situations, $C_g$ is a number-based self-adaptive parameter.

**3.4 Constrained nondominated-solution selection**

In multi-objective algorithms, nondominated solutions are selected according to the definition of the nondominated solution whose fitness is no worse than the others.

However, most solutions generated in the early stage of the evolutionary procedure are poor compared with the solutions obtained in the final generation. The most significant problem is that the nondominated solutions are selected from these poor solutions simply according to the definition of the nondominated solution in the early generations. Hence, to increase the likelihood of obtaining good nondominated solutions, the fitness of each nondominated solution must satisfy the following constraints to prevent the foregoing problem.

$$F_R(X_i) \geq 0.1 \text{ and } F_C(X_i) \leq 100000. \tag{17}$$

### 3.5 Crowding distance and repository

Similar to MOPSO, there is a size-limited repository $N_{rep}$ in the proposed MOSSO to store all the found local nondominated solutions. If the number of local nondominated solutions ($N_{lns}$) is larger than $N_{rep}$, these local nondominated solutions with smaller crowding distances are discarded until the number of $N_{lns}$ does not exceed that of $N_{rep}$. The concepts of the repository and the crowding distance are adopted from MOPSO [44] and NSGA-II [10, 14-16], respectively, and are combined to utilize the advantages of both in the MOSSO.

### 3.6 New *pBest* replacement policy

If the new updated solution $X_i$ dominates its *pBest* $P_i$, $P_i$ is replaced with $X_i$ (and vice versa) in both MOPSO and MOSSO. However, in MOPSO, either $X_i$ or $P_i$ is picked randomly to be the new *pBest* and is added to the repository if they are not dominated by each other. Therefore, if the old $P_i$ is added to the repository, the MOPSO requires additional time to verify whether such a $P_i$ is a local nondominated solution, because $P_i$ had been tested in the first generation when it was a *pBest*. Hence, MOPSO may lose a global nondominated solution if $X_i$ is not dominated by $P_i$ and $X_i$ is not replaced $P_i$ for $i = 1, 2, \ldots, N_{sol}$.

To fix this problem in the MOPSO, $X_i$ is always added to the repository, and the old $P_i$ is replaced with the new $P_i$ if $X_i$ and the old $P_i$ are not dominated by each other in the MOSSO for $i = 1, 2, \ldots, N_{sol}$.

**3.7 Full-factorial design**

Several different versions of SSO have been proposed for numerous applications [5, 7, 8, 25-43]. Three major modifications are discussed below.

1. Compulsory replacement [5, 8, 25-37, 39, 41-43] vs. survival-of-the-fittest [7, 38, 40]

    Let the solution $X_{new}$ be updated from the solution $X_{old}$. In almost all AI algorithms, $X_{new}$ replaces $X_{old}$ if $F(X_{new})$ is better than $F(X_{old})$. However, there are two different strategies to determine whether to keep or discard $X_{new}$ or $X_{old}$ if $F(X_{new})$ is worse than $F(X_{old})$: compulsory replacement [5, 8, 25-37, 39, 41-43] and survival-of-the-fittest [7, 38, 40]. In the former, $X_{new}$ must replace $X_{old}$ regardless of whether $F(X_{new})$ is worse than $F(X_{old})$ [5, 8, 25-37, 39, 41-43]. In the latter, $X_{old}$ is kept, and $X_{new}$ is discarded [7, 38, 40].

2. All-variable update [5, 7, 8, 25-27, 29-38, 40-43] vs. one-variable update [28, 39]

    The all-variable update first used in the traditional SSO allows each variable in any solution to be updated at its own pace [5, 7, 8, 25-27, 29-38, 40-43]. Hence, the all-variable update is completely different from the vectorized update mechanism in PSO [9, 11, 12] and the crossover update mechanism in GA [10, 19]. The all-variable update provides more energy to update solutions to pass over the local traps [5, 7, 8, 25-27, 29-38, 40-43]. If the problem requires a global search, the all-variable update is the best choice. In contrast, in the one-variable update mechanism, only one randomly selected variable in each solution is updated; the other variables are not changed [28, 39]. Hence, the one-variable update is suitable for problems without a large number of local optima.

    Both update mechanisms have strengths and weaknesses [5, 7, 8, 25-43]. The one-variable update process converges to the local optima faster than the all-variable update but may be trapped in local optima. The all-variable update process can easily to escape from the local optimum, but it may take more time to reach the real optimum, even if the current solution is very close to the real optimum.

3. With *pBest* [5, 7, 8, 25-29, 31-40, 43] vs. without *pBest* [29, 41, 42]

The *pBest* of a solution, e.g., $X_i$, is the best solution in the update process of the solution thus far. In PSO [9, 11, 12] and the traditional SSO [5, 7, 8, 25-29, 31-40, 43], changing *pBest* can prevent *X* from being updated to a poor solution in the next generation. However, in various cases, the "without *pBest*" approach is a better choice [29, 41, 42].

To efficiently and systematically determine which of the aforementioned versions of SSO should be included in the proposed MOSSO for optimizing the performance, a full-factorial design (design of experiments method) is implemented by treating the three aforementioned modifications as the three factors. Each factor has two levels, yielding eight different versions of MOSSO, as shown in Table 3.

Table 3. Full-factorial design.

| Factor \ Level | 0 | 1 |
|---|---|---|
| 1 | compulsory replacement | survival-of-the-fittest |
| 2 | all-variable update | one-variable update |
| 3 | with *pBest* | without *pBest* |

The notation XXX is used in the name of MOSSO, where X ∈ {0, 1}, and the first, second, and third digits indicate the levels of the first, second, and third factors, respectively. The factors and levels for the eight versions of MOSSO are presented in Table 4.

Table 4. Eight versions of MOSSO.

| MOSSO \ Factor | 1 | 2 | 3 |
|---|---|---|---|
| MOSSO-000 | compulsory replace | all-variable update | with *pBest* |
| MOSSO-001 | compulsory replace | all-variable update | no *pBest* |
| MOSSO-010 | compulsory replace | one-variable update | with *pBest* |
| MOSSO-011 | compulsory replace | one-variable update | no *pBest* |
| MOSSO-100 | survival-of-the-fittest | all-variable update | with *pBest* |
| MOSSO-101 | survival-of-the-fittest | all-variable update | no *pBest* |
| MOSSO-110 | survival-of-the-fittest | one-variable update | with *pBest* |
| MOSSO-111 | survival-of-the-fittest | one-variable update | no *pBest* |

For example, "MOSSO-001" indicates that compulsory replace, all-variable update, no *pBest* are used in the MOSSO.

### 3.8 Complete procedure of proposed MOSSO

The procedure of the proposed MOSSO-000 method is described in this section. Accordingly, the procedures for the other seven versions of MOSSO, i.e., MOSSO-001, MOSSO-010, …, and MOSSO-111, can be developed on the basis of Section 3.7.

**MOSSO-000 PROCEDURE**

**STEP 0.** Generate solutions $P_1 = X_1$, $P_2 = X_2$, …, $P_{N_{sol}} = X_{N_{sol}}$ randomly, calculate the fitness functions of each objective, let $t = i = 1$, and set the repository $\Omega = $ {all local nondominated solutions $X_i$ with $F_R(X_i) \geq 0.1$ and $F_C(X_i) \leq 100000$}.

**STEP 1.** Update $X_i$ based on Eq. (16) and calculate its fitness values for all objective functions based on Eq. (1).

**STEP 2.** If $P_i$ is dominated $X_i$, proceed to STEP 5. Otherwise, let $P_i = X_i$.

**STEP 3.** If $F_R(X_i) < 0.1$ or $F_C(X_i) > 100000$, proceed to STEP 5.

**STEP 4.** $\Omega = $ {all local nondominated solutions in $\Omega \cup \{X_i\}$}.

**STEP 5.** Let $i = i + 1$ and proceed to STEP 1 if $i < N_{sol}$.

**STEP 6.** If $|\Omega| > N_{rep}$, update $\Omega$ by selecting the $N_{rep}$ solutions with the largest crowding distances in $\Omega$.

**STEP 7.** Let $t = t + 1$ and $i = 1$, and proceed to STEP 1 if $t < N_{gen}$. Otherwise, halt, and $\Omega$ is the local Pareto front.

## 4. NUMERICAL EXAMPLES

All 10 algorithms, including the eight different versions of MOSSO based on the full-factorial design discussed in Section 3.7, MOPSO, and NSGAII, are coded in Dev C++5.12, run on an Intel Core i7 3.07-GHz personal computer with 16 GB of memory, and measured with regard to the runtime based on Central Processing Unit (CPU) seconds. Each algorithm is executed 50 times (i.e., $N_{run} = 50$) independently with a solution number of $N_{sol} = 100$ and a generation number of $N_{gen} = 1000$, which is also the stopping criterion. Moreover, to obtain realistic solutions, there is a lower bound for all the reliability variables: $R_{lb} = 0.75$.

The number of local nondominated solutions in the repository is equal to the number of solutions, i.e., $N_{sol} = N_{rep}$, in MOPSO and these MOSSO that used repository. In NSGA-II, one-cut crossover and mutation are implemented to update the solutions. In contrast, the traditional update equation used in

PSO is adopted to update both types of solutions in MOPSO:

$$V_{\mathbf{n},i} = \text{Min}\{\text{Max}\{w \times V_{\mathbf{n},i} + c_1 \times \rho_1 \times (P_{\mathbf{n},i} - \mathbf{n}_i) + c_2 \times \rho_2 \times (P_{\mathbf{n},gBest} - \mathbf{n}_i), V_{\mathbf{n},\min}\}, V_{\mathbf{n},\max}\} \quad (18)$$

$$\mathbf{n}_i = \text{Min}\{\text{Max}\{\mathbf{n}_i + V_{\mathbf{n},i}, \mathbf{n}_{\min}\}, \mathbf{n}_{\max}\} \quad (19)$$

$$V_{\mathbf{r},i} = \text{Min}\{\text{Max}\{w \times V_{\mathbf{r},i} + c_1 \times \rho_1 \times (P_{\mathbf{r},i} - \mathbf{r}_i) + c_2 \times \rho_2 \times (P_{\mathbf{r},gBest} - \mathbf{r}_i), V_{\mathbf{r},\min}\}, V_{\mathbf{r},\max}\} \quad (20)$$

$$\mathbf{r}_i = \text{Min}\{\text{Max}\{\mathbf{r}_i + V_{\mathbf{r},i}, \mathbf{r}_{\min}\}, \mathbf{r}_{\max}\} \quad (21)$$

The other required parameters for all the algorithms are listed below:

MOSSO: $C_g = 0.7$ and $C_w = 0.9$ if no *pBest* is used; $C_g = 0.5$, $C_p = 0.75$, and $C_w = 0.9$.

MOPSO: $w = .5$, $c_1 = c_2 = 0.5$, $V_{\mathbf{n},\max} = V_{\mathbf{r},\max} = 0.5$, $V_{\mathbf{n},\min} = V_{\mathbf{r},\min} = -0.5$, $n_{\min} = 1$, $n_{\max} = 10$, $r_{\min} = 0.5$, and $r_{\max} = 1$.

NSGA2: The crossover and mutation rates are 0.6 and 0.4, separately, such that the numbers of new solutions generated from the crossover operation and mutation are 60 and 40, respectively.

The experimental results are presented in Tables 5–10 and Figures 1–4 and are analyzed in the following subsections.

**4.1 Performance comparison based on convergence or diversity**

As mentioned in Section 2.2, most indices for measuring the performance of multi-objective algorithms calculate either the convergence or the diversity between local nondominated solutions and the global Pareto front. Among these popular measure indices, the Generational distance (GD) [48] and spacing (SP) [49] are the most widely used for calculating the convergence index and diversity index, respectively.

Let $d_i$ be the distance between the $i$th local nondominated solution $X_i$ and its nearest neighbor in the Pareto front, $\phi_{k,j}$ be the $j$th objective value of the $k$th solution in the Pareto front, and $\bar{d}$ be the average sum of all $d_i$, where $i = 1, 2, ..., N_{lns}$, $j = 1, 2, ..., N_{obj}$, and $k = 1, 2, ..., N_{gns}$:

$$d_i = \min\left\{ \sqrt{\sum_{i=1}^{N_{lns}} (f_{i,j} - \phi_{k,j})^2} \;\middle|\; \text{for } k = 1, 2, ..., N_{gns} \right\} \tag{22}$$

$$\bar{d} = \frac{\sum_{i=1}^{N_{lns}} d_i}{N_{lns}} \tag{23}$$

The GD is defined as the average Euclidean distances between the local nondominated solutions and the global Pareto front, as shown below:

$$GD = \frac{\sqrt{\sum_{i=1}^{N_{lns}} d_i^2}}{N_{lns}}. \tag{24}$$

In contrast to the GD, which represents the average Euclidean distances, the SP indicates the diversity of the local nondominated solutions along the Pareto front. The SP is similar to the standard deviation and can be expressed as follows [49]:

$$SP = \sqrt{\frac{\sum_{i=1}^{N_{lns}} (\bar{d} - d_i)^2}{N_{lns} - 1}}. \tag{25}$$

In general, a shorter GD indicates that the solution is closer to the Pareto front, and a larger SP corresponds to better diversity of the found nondominated solutions.

The value of $d_i$ is the basis of all the measurement indices, including the GD and SP, from Eqs. (24) and (25). To solve Eq. (23), the Pareto front is required; however, it is impossible to form a complete Pareto front, because a curve contains an infinite number of points; i.e., a front includes infinite global nondominated solutions. Selecting nondominated solutions from local nondominated solutions obtained in the last generation to simulate the Pareto front, which is called a "simulated Pareto front," is a common way to overcome the aforementioned obstacle.

Hence, in our experiments, all the local nondominated solutions obtained from the 10 algorithms are collected, and these nondominated solutions filtered out from all the local nondominated solutions in the collection are used to simulate the Pareto front. Table 5 presents the number of found local nondominated solutions $N_{lns}$, the number of nondominated solutions $N_{gns}$ in the simulated Pareto front, the number of infeasible solutions $N_{inf}$, and the GD and SP values. The bold and underlined values represent the best and the second-best among all the related values, respectively.

**Table 5.** Summary of the experimental results

| ID | Algorithm | $N_{lns}$ | $N_{gns}$ | $N_{inf}$ | GD | SP |
|---|---|---|---|---|---|---|
| 1 | MOSSO-000 | 4947 | 226 | 53 | 0.0025486834 | 0.0358543545 |
|   | MOSSO-001 | 4954 | <u>280</u> | 46 | 0.0025788064 | 0.0379025526 |
|   | MOSSO-010 | <u>4967</u> | 78 | <u>23</u> | <u>0.0024322215</u> | 0.0283098072 |
|   | MOSSO-011 | 4948 | 97 | 51 | 0.0024709394 | 0.0304954536 |
|   | MOSSO-100 | 4962 | 274 | 38 | 0.0025610263 | 0.0363749228 |
|   | MOSSO-101 | 4953 | 253 | 47 | 0.0025913157 | 0.0372196548 |
|   | MOSSO-110 | 4946 | 61 | 39 | 0.0024418700 | 0.0298935752 |
|   | MOSSO-111 | 4963 | 65 | 37 | 0.0024829882 | 0.0307577625 |
|   | MOPSO | 55 | 0 | 4775 | 0.0555312932 | <u>0.0193608515</u> |
|   | NSGA-II | **5000** | **338** | **0** | **0.0008441776** | **0.0121382810** |
|   | Total | 44695 | 1672 | 5109 | | |
| 2 | MOSSO-000 | 4990 | 167 | 10 | 0.0035900876 | 0.1013769135 |
|   | MOSSO-001 | 4983 | 233 | 17 | 0.0036576760 | 0.1050197035 |
|   | MOSSO-010 | 4992 | 125 | 5 | 0.0032849144 | 0.0728718713 |
|   | MOSSO-011 | 4992 | 95 | 5 | 0.0030304755 | 0.0716783106 |
|   | MOSSO-100 | 4990 | 169 | 10 | 0.0035877863 | 0.1000497267 |
|   | MOSSO-101 | 4983 | <u>239</u> | 17 | 0.0035901975 | 0.1027649119 |
|   | MOSSO-110 | 4989 | 77 | 10 | 0.0033958696 | 0.0769449025 |
|   | MOSSO-111 | <u>4997</u> | 81 | <u>3</u> | <u>0.0029895080</u> | <u>0.0705384612</u> |
|   | MOPSO | 2271 | 0 | 2452 | 0.0062204800 | 0.0743392184 |
|   | NSGA-II | **5000** | **240** | **0** | **0.0019613253** | **0.0329662710** |
|   | Total | 47187 | 1426 | 2529 | | |
| 3 | MOSSO-000 | 4987 | 208 | 13 | 0.0030357849 | 0.0607248098 |
|   | MOSSO-001 | 4972 | 217 | 28 | 0.0030174130 | 0.0617981590 |
|   | MOSSO-010 | <u>4995</u> | 82 | <u>4</u> | 0.0032460494 | 0.0544829741 |
|   | MOSSO-011 | 4989 | 34 | 9 | <u>0.0029362889</u> | <u>0.0474426486</u> |
|   | MOSSO-100 | 4985 | <u>254</u> | 15 | 0.0029893676 | 0.0599902011 |
|   | MOSSO-101 | 4984 | **271** | 16 | 0.0030124336 | 0.0605894327 |
|   | MOSSO-110 | 4992 | 93 | <u>4</u> | 0.0031577374 | 0.0506542362 |
|   | MOSSO-111 | 4993 | 52 | 6 | 0.0029823731 | 0.0503170006 |
|   | MOPSO | 2857 | 0 | 2039 | 0.0054357145 | 0.0581620410 |
|   | NSGA-II | **5000** | 207 | **0** | **0.0015246626** | **0.0133272447** |
|   | Total | 47754 | 1418 | 2134 | | |
| 4 | MOSSO-000 | 3879 | <u>199</u> | 1121 | 0.0051039513 | 0.1153833643 |
|   | MOSSO-001 | 3817 | 187 | 1183 | 0.0052134302 | 0.1201714352 |
|   | MOSSO-010 | 4128 | 165 | 870 | 0.0046740957 | 0.1022532582 |
|   | MOSSO-011 | <u>4277</u> | 148 | <u>723</u> | 0.0046561025 | 0.1043761075 |
|   | MOSSO-100 | 3851 | 152 | 1149 | 0.0052661696 | 0.1228384674 |
|   | MOSSO-101 | 3801 | 174 | 1199 | <u>0.0052893702</u> | 0.1216250509 |
|   | MOSSO-110 | 4046 | 132 | 952 | 0.0046892711 | <u>0.1012461036</u> |
|   | MOSSO-111 | 4247 | 162 | 750 | 0.0046848543 | 0.1087769419 |
|   | MOPSO | 0 | 0 | 4761 | | |
|   | NSGA-II | **4998** | **479** | 2 | **0.0019406326** | **0.0546758622** |
|   | Total | 37044 | 1798 | 12710 | | |

As shown in Table 5, NSGA-II is almost the best among the 10 algorithms tested on the four benchmark problems with regard to $N_{lns}$, $N_{gns}$ (except for ID = 3), $N_{inf}$, GD, and SP. However, a further analysis based on the plots reveals that the GD and SP are insufficient for determining which algorithm is better; i.e., NSGA-II is not as good as Table 5 indicates, as discussed in Sections 4.2 and 4.3.

Unexpectedly, the MOSPSO appears to be unsuitable for bi-objective active RRAP problems and has zero nondominated solutions. This is because the MOPSO is based on PSO, and the handling of discrete variables is a weakness of PSO. However, it is difficult to explain why the number of obtained local nondominated solutions is zero and all the solutions in the repository are infeasible.

The information in Table 5 is insufficient to determine the best version of MOSSO among the eight versions. Hence, further analysis is performed, as described in Sections 4.2 and 4.3.

**4.2 Plots of simulated Pareto fronts**

The individual simulated Pareto fronts for all 10 algorithms (as shown in Table 6), i.e., the eight versions of MOSSO (Labels 0–7), MOSPSO (Label 8), and NSGA-II (Label 9), are presented in Fig. 1–4. The infeasible solutions and all the local nondominated solutions in the simulated Pareto front are removed. The titles above the plots indicate the algorithms.

**Table 6.** Labels and corresponding algorithms.

| Label | Algorithm | Label | Algorithm |
|---|---|---|---|
| 0 | MOSSO-000 | 5 | MOSSO-101 |
| 1 | MOSSO-001 | 6 | MOSSO-110 |
| 2 | MOSSO-010 | 7 | MOSSO-111 |
| 3 | MOSSO-011 | 8 | MOPSO |
| 4 | MOSSO-100 | 9 | NSGA-II |

These Pareto plots for MOPSO and NSGA-II confirm our observations presented in Section 4.1. The MOSPSO is the worst algorithm, with no simulated Pareto front for any of the four benchmark RRAPs. This is explained in Section 4.1. The Pareto fronts of NSGA-II are all gathered in the upper parts of the Pareto fronts. This explains why NSGA-II always has the best GD among the 10 algorithms and indicates that the GD is insufficient to determine which multi-objective algorithm has the best performance.

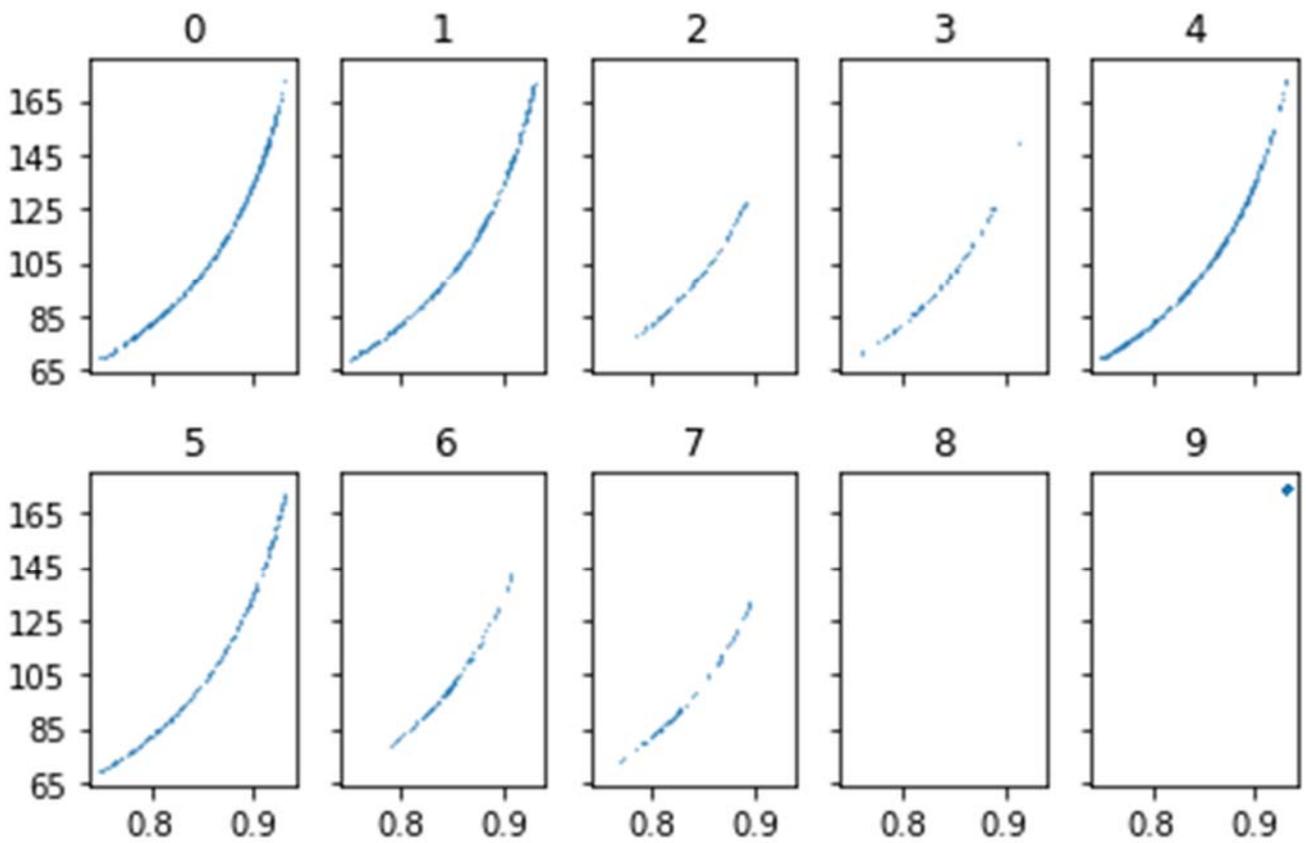

**Figure 1.** Individual simulated Pareto front for Benchmark 1.

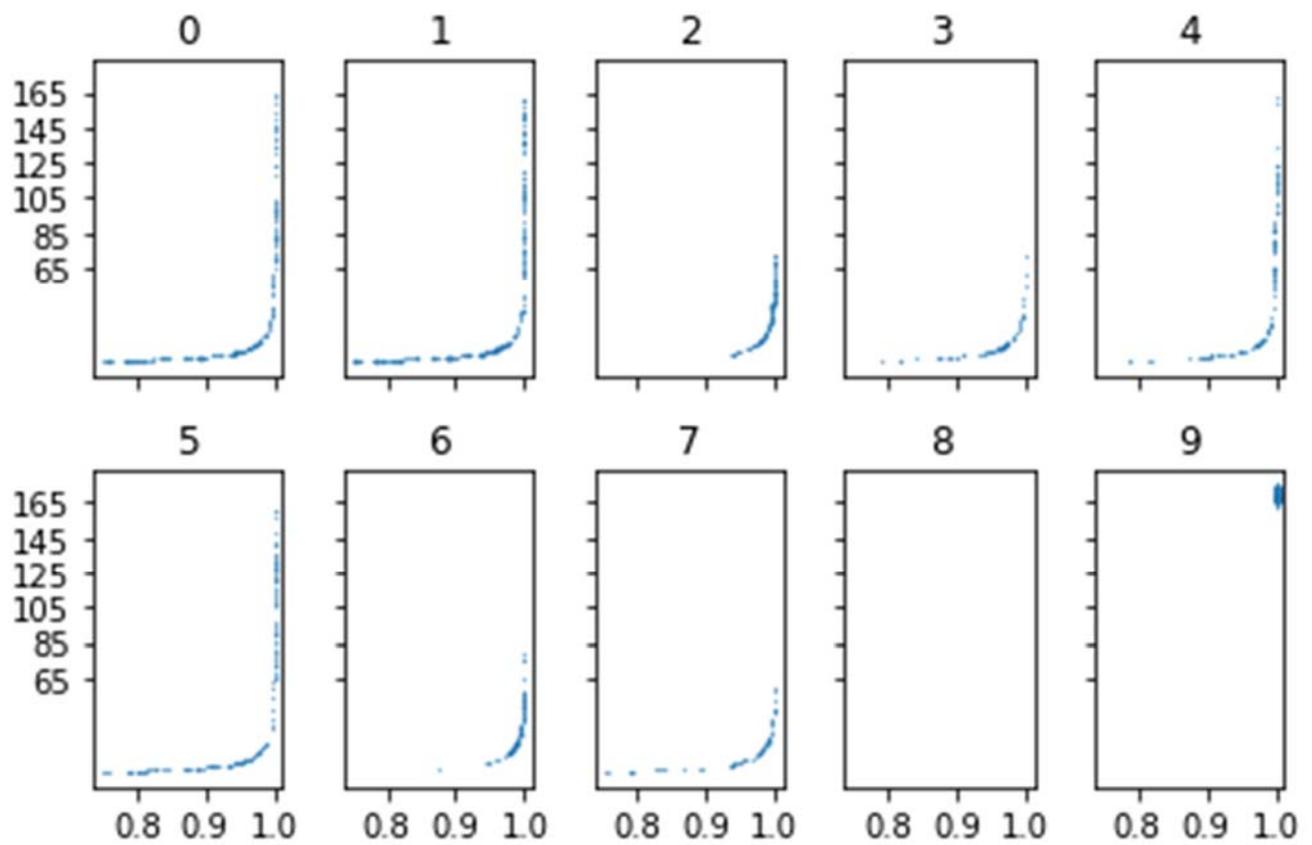

**Figure 2.** Individual simulated Pareto front for Benchmark 2.

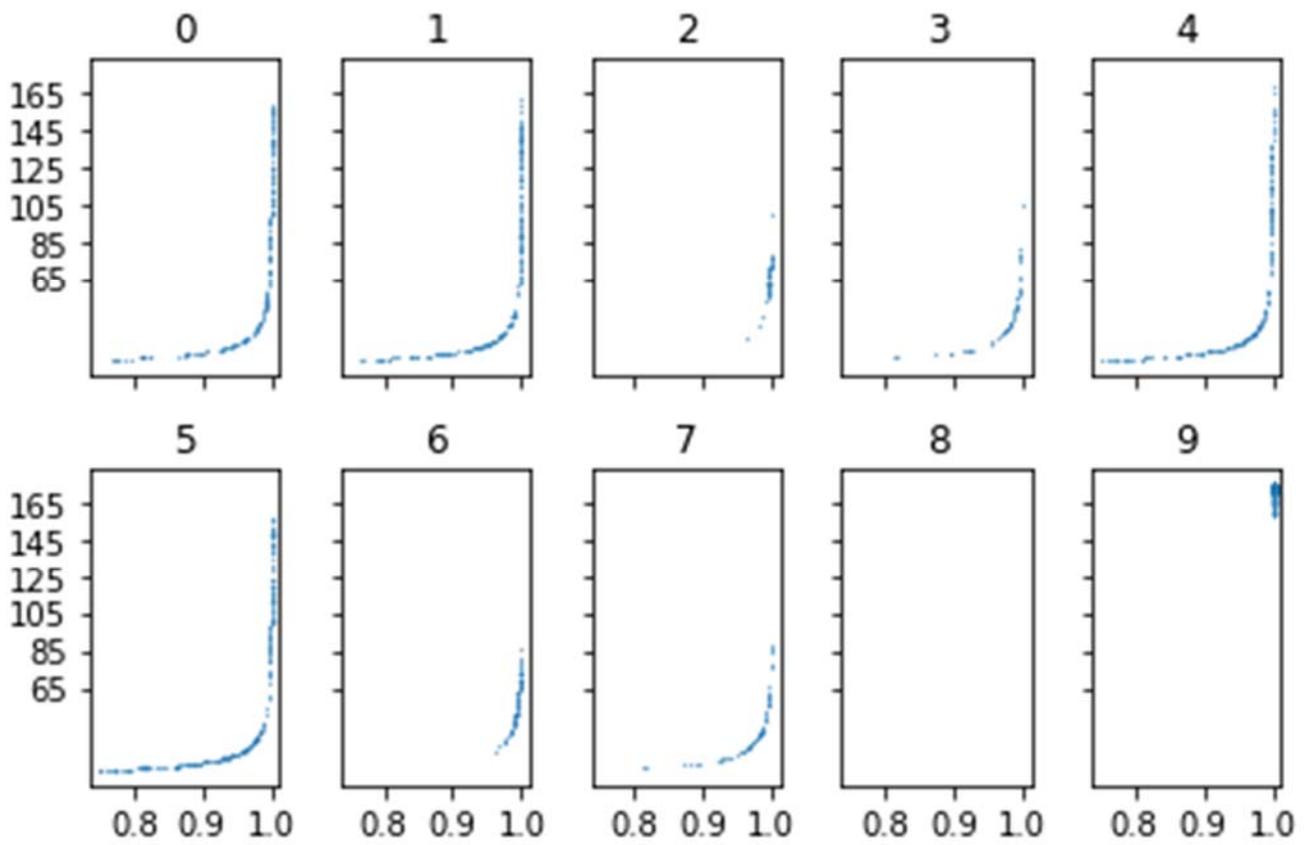

**Figure 3.** Individual simulated Pareto front for Benchmark 3.

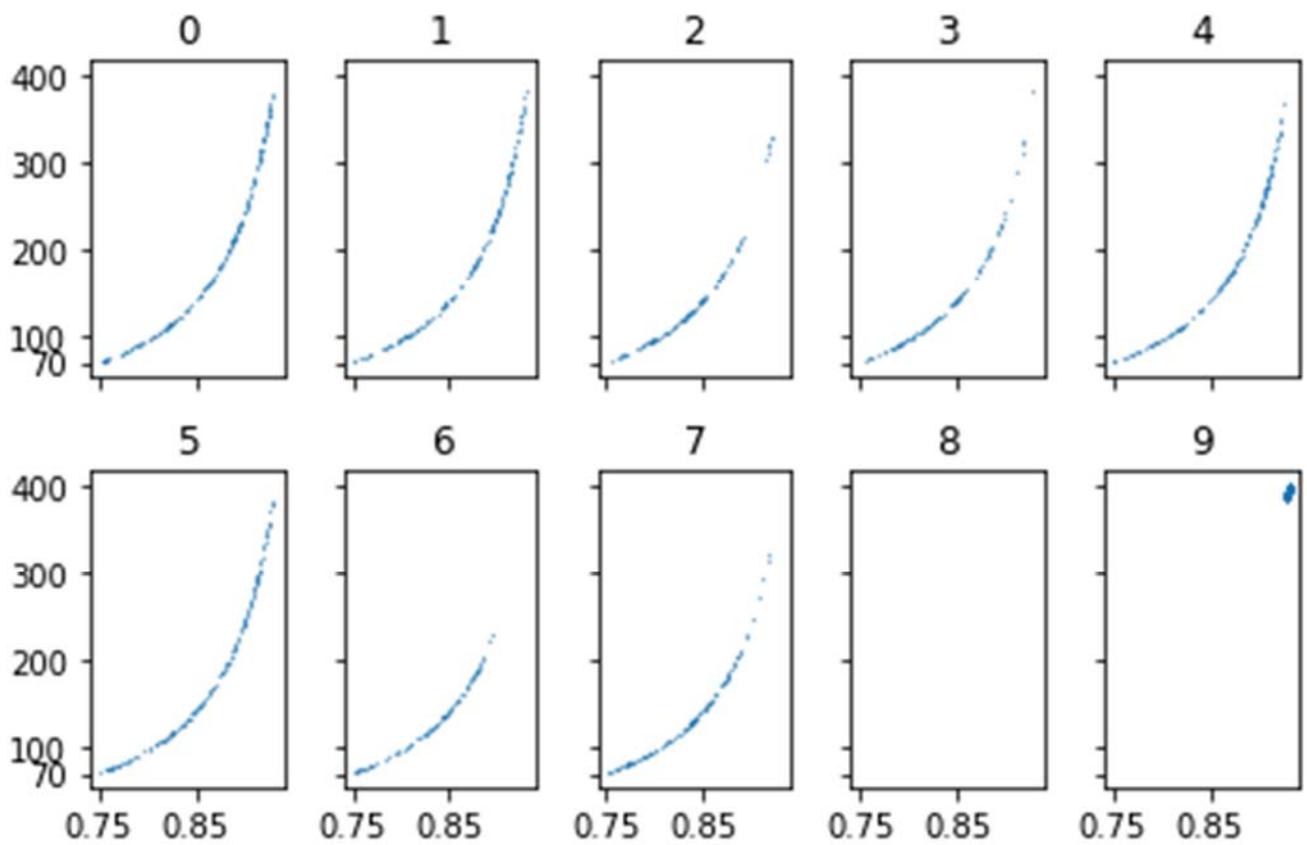

**Figure 4.** Individual simulated Pareto front for Benchmark 4.

The reasons why NSGA-II exhibits such peculiar phenomena (good with regard to the convergence index but poor in reality) are presented below:

1. The elite selection always selects better solutions for the next generation. However, the discarded solutions may become better than the solutions selected in the later generations. Hence, the elite selection has a high probability of avoiding new solutions selected for the next generation.

2. The nondominated sorting, which is used in STEP G2 of Section 2.5, fails to maintain the diversity of local nondominated solutions if the number of local nondominated solutions gathered together is greater than or equal to $N_{sol}$ in Rank 1; i.e., no solutions are selected with other rankings.

3. The versions of MOSSO with one-variable update, i.e., MOSSO-010, MOSSO-011, MOSSO-110, and MOSSO-111, all have poor $N_{lns}$, $N_{gns}$, $N_{inf}$, and GD results compared with those without one-variable update. Section 4.3 provides evidence that factor 2 with level 1 (e.g., the all-variable update) represents the optimal version of MOSSO. The mutation used in NSGA-II is identical to the one-variable update used in MOSSO, except that it is not implemented fully. Hence, mutation is not suitable for the bi-objective cold RRAP.

4. A mutation introduces a new change to a variable even it is not helpful, as explained previously. In contrast, a crossover only exchanges information that is already known between two parents to generate offspring, and such close inbreeding increases the probability of local nondominated solutions gathering together.

**Table 7.** Two best groups according to the local Pareto fronts.

| ID \ Group | 1 | 2 |
|---|---|---|
| 1 | 1 | 0, 5 |
| 2 | 1 | 0, 5 |
| 3 | 1 | 0, 4, 5 |
| 4 | 0, 1, 5 | 4 |

Hence, according to the plots, the MOSSO outperforms NSGA-II and MOPSO. However, it is difficult to determine which MOSSO is the best visually. Hence, we define two groups: Group 1 (algorithms with better Pareto fronts, i.e., those with a larger gap than the others) and Group 2. Any algorithm in Group 1 is better than all the algorithms in Group 2; however, the Pareto fronts in the same group are almost identical across the different algorithms. Algorithms not listed in these two groups are not worthy of discussion, because there is a large gap between them and the two groups.

For example, these Pareto fronts with Labels 0, 1, and 5 are better than the other 7 Pareto fronts, and these three Pareto fronts are separated further into Groups 1 and 2. According to our rules, the groups for the 10 algorithms and four benchmark RRAPs are presented below.

As shown in Table 7, Labels 0, 1, and 5 are always listed in the two groups for each ID. Additionally, Label = 1 is always in Group 1. Therefore, according to the full-factorial design and Table 7, the best MOSSO is with compulsory replacement, all-variable update, and without *pBest*, i.e., MOSSO-001 (Label 1).

### 4.3 Finding optimal MOSSO

According to the plots in Section 4.2, all eight versions of MOSSO outperform NSGA-II and MOPSO. A further analysis based on the full-factorial design for selecting the optimal MOSSO is presented here.

The results obtained for the eight versions of MOSSO are presented in Table 8. The gap represents the percentage difference between two distinctive levels for the same factor and index. The gap indicates which factor is improved the most (if its level is changed). For example, the value 0.04% under factor 1 in $N_{lns}$ for ID = 1 indicates that the $N_{lns}$ value is improved by 0.04% if the level is changed from 0 to 1. This is expressed as follows:

$$\frac{|\text{better}\{4954, 4956\} - \text{worse}\{4954, 4956\}|}{\text{better}\{4954, 4956\}} = 0.04\%. \tag{19}$$

The bold font indicates that the level value is better than that of another level for the same factor, index, and ID. For example, 4956 is in bold because it is better than 4954 for $N_{lns}$ columns, Factor = 1, and ID = 1. For the gap, the bold font indicates that the value is better than those for the other factors; e.g., the gap value 70.9% is bold because it is the best among 4.1%, 70.9%, and 8.1% for $N_{gns}$ columns and ID = 1.

According to the gap values in Table 8, Factor = 2 (for which Level = 0 corresponds to the all-variable update and Level = 1 corresponds to the one-variable update) is the most important among the three factors, having better values of $N_{lns}$, $N_{gns}$, $N_{inf}$, GD, and SP. The gap values reach ≥ 70.9% in $N_{gns}$ for IDs

= 1 and 3; ≥ 37.8% in $N_{inf}$ for ID = 2 and 3; ≥ 10.39% in GD for ID = 2 and 4; and ≥ 13% in the SP for all the IDs. Moreover, Level = 0 is always better than Level = 1 for Factor = 2 in most parts. Thus, for these four benchmark RRAPs, Level = 0, i.e., all-variable update, with Factor = 2 is the best version of MOSSO to ensure good convergence and diversity. For Level = 0 and Factor = 2, it is more important to have a strong global search ability than a strong local search ability for the proposed RRAP.

For Factors 1 and 3, it is not as easy as Factor 2 to find the best level for all IDs. Roughly, Level 0 is better than Level 1 for IDs 2 and 3, but Level 1 outperforms Level 0 for IDs 1 and 4. To select the levels of the other two factors, Factor 2 is removed completely from Table 8, and the corresponding results are presented in Table 9. As shown, Level = 1 always has better values than Level = 0 for Factor = 3, except for the values of $N_{lns}$ (for which the value of Level 1 is only 0.105% worse than that of Level 0 for IDs = 1, 2, 3, but 1.47% better than that of Level 0 for ID = 4; i.e., the difference in $N_{lns}$ is negligible). Hence, Level = 1 is the better option for Factor =3 in the proposed MOSSO.

Table 8. Summary of the eight MOSSOs according to the levels of each factor.

| ID | Level\Factor | $N_{lns}$ 1 | 2 | 3 | $N_{gns}$ 1 | 2 | 3 | $N_{inf}$ 1 | 2 | 3 | GD 1 | 2 | 3 | SP 1 | 2 | 3 |
|---|---|---|---|---|---|---|---|---|---|---|---|---|---|---|---|---|
| 1 | 0 | 4954 | 4954 | **4955.5** | 170.3 | **258.3** | 159.8 | **43.3** | **43.3** | 38.3 | 0.002508 | **0.002570** | 0.002496 | 0.033141 | **0.036838** | 0.032608 |
|   | 1 | **4956** | **4956** | 4954.5 | 163.3 | 75.3 | **173.8** | 40.3 | 37.5 | **45.3** | **0.002519** | 0.002457 | **0.002531** | **0.033561** | 0.029864 | **0.034094** |
|   | gap% | 0.04% | **0.04%** | 0.02% | 4.1% | **70.9%** | 8.1% | 6.9% | 13.3% | **15.5%** | 0.462% | **4.395%** | 1.385% | 1.254% | **18.931%** | 4.358% |
| 2 | 0 | 4989.3 | 4986.5 | 4990.3 | **155.0** | **202.0** | 134.5 | 9.3 | **9.3** | 8.8 | 0.003391 | **0.003606** | **0.003465** | **0.087737** | **0.102303** | **0.087811** |
|   | 1 | **4989.8** | **4992.5** | 4988.8 | 141.5 | 94.5 | **162.0** | **10.0** | 5.8 | **10.5** | **0.003391** | 0.003175 | 0.003317 | 0.087575 | 0.073008 | 0.087500 |
|   | gap | 0.01% | **0.12%** | 0.03% | 8.7% | **53.2%** | 17.0% | 7.5% | **37.8%** | 16.7% | 0.002% | **11.958%** | 4.263% | 0.185% | **28.635%** | 0.354% |
| 3 | 0 | 4985.8 | 4982.0 | **4989.8** | 135.3 | **237.5** | 159.3 | **13.5** | **13.5** | 9.0 | **0.003059** | 0.003014 | **0.003107** | **0.056112** | **0.060776** | **0.056463** |
|   | 1 | **4988.5** | **4992.3** | 4984.5 | **167.5** | 65.3 | 143.5 | 10.3 | 5.8 | **14.8** | 0.003035 | **0.003081** | 0.002987 | 0.055388 | 0.050724 | 0.055037 |
|   | gap | 0.055% | **0.205%** | 0.105% | 19.3% | **72.5%** | 9.9% | 24.1% | **57.4%** | 39.0% | 0.765% | 2.170% | **3.865%** | 1.291% | **16.539%** | 2.526% |
| 4 | 0 | **4025.3** | 3837.0 | 3976.0 | **174.8** | **178.0** | 162.0 | 974.3 | **974.3** | **1023.0** | 0.004912 | **0.005218** | 0.004933 | 0.110546 | **0.120005** | 0.110430 |
|   | 1 | 3986.3 | **4174.5** | **4035.5** | 155.0 | 151.8 | **167.8** | **1012.5** | 823.8 | 963.8 | **0.004982** | 0.004676 | **0.004961** | **0.113622** | 0.104163 | **0.113737** |
|   | gap | 0.97% | **8.08%** | 1.47% | 11.3% | **14.7%** | 3.4% | 3.8% | **15.4%** | 5.8% | 1.415% | **10.390%** | 0.556% | 2.707% | **13.201%** | 2.908% |

Table 9. Summary of the eight MOSSOs after the removal of Factor 2.

| ID |   | $N_{lns}$ 1 | 3 | $N_{gns}$ 1 | 3 | $N_{inf}$ 1 | 3 | GD 1 | 3 | SP 1 | 3 |
|---|---|---|---|---|---|---|---|---|---|---|---|
| 1 | 0 | 4950.5 | **4954.5** | 253.0 | 250.0 | **49.5** | 45.5 | 0.002564 | 0.002555 | **0.036878** | 0.036115 |
|   | 1 | 4957.5 | 4953.5 | **263.5** | **266.5** | 42.5 | **46.5** | **0.002576** | **0.002585** | 0.036797 | **0.037561** |
|   | gap | **0.14%** | 0.02% | 4.0% | **6.2%** | **14.1%** | 2.2% | 0.482% | **1.168%** | 0.220% | **3.851%** |
| 2 | 0 | 4986.5 | **4990.0** | 200.0 | 168.0 | **13.5** | 10.0 | **0.003624** | 0.003589 | **0.103198** | 0.100713 |
|   | 1 | 4986.5 | 4983.0 | **204.0** | **236.0** | **13.5** | **17.0** | 0.003589 | **0.003624** | 0.101407 | **0.103892** |
|   | gap | 0.00% | **0.14%** | 2.0% | **28.8%** | 0.0% | **41.2%** | 0.963% | 0.966% | 1.735% | **3.060%** |
| 3 | 0 | 4979.5 | **4986.0** | 212.5 | 231.0 | **20.5** | 14.0 | **0.003027** | 0.003013 | **0.061261** | 0.060358 |
|   | 1 | **4984.5** | 4978.0 | **262.5** | **244.0** | 15.5 | **22.0** | 0.003001 | **0.003015** | 0.060290 | **0.061194** |
|   | gap | 0.100% | **0.160%** | **19.0%** | 5.3% | 24.4% | **36.4%** | **0.849%** | 0.078% | **1.586%** | 1.367% |
| 4 | 0 | **3848.0** | **3865.0** | **193.0** | 175.5 | 1152.0 | 1135.0 | 0.005159 | 0.005185 | 0.117777 | 0.119111 |
|   | 1 | 3826.0 | 3809.0 | 163.0 | **180.5** | **1174.0** | **1191.0** | **0.005278** | **0.005251** | **0.122232** | **0.120898** |
|   | gap | 0.57% | **1.45%** | **15.5%** | 2.8% | 1.9% | **4.7%** | **2.256%** | 1.263% | **3.644%** | 1.478% |

For Factor 1, both Levels 0 and 1 have nine better values and two values in tie from Table 10. Level 0 has more numbers of gap values that are >10%. Hence, there is no conclusive evidence to determine which level is better. However, Level 0 is a better choice if we must pick one from Levels 0 and 1.

Table 10. Summary of Factor = 1 based on the gap values.

| Gap | Level 0 | Level 1 |
| --- | --- | --- |
| (20% - 25%] | 1 ($N_{inf}$, ID=3) | 0 |
| (15% - 20%] | 1 ($N_{gns}$, ID=4) | 1 ($N_{gns}$, ID=3) |
| (10% - 15%] | 1 ($N_{inf}$, ID=1) | 0 |
| (5% - 10%] | 0 | 0 |
| (1% - 5%] | 2 | 5 |
| (0.5% - 1%] | 3 | |
| (0% - 0.5%] | 1 | 3 |
| Total | 9 | 9 |

The conclusion of Levels = 0, 0, 1 for Factors 1–3, respectively, match the observations regarding MOSSO presented in last paragraph of Section 4.2.

# 5. CONCLUSIONS

A new reliable design problem is proposed to reduce the cost and enhance the reliability of solutions in active RRAP for real-life scenarios. A multi-objective SSO called the MOSSO is customized and developed to solve this bi-objective active RRAP problem.

The MOSSO employs the one-type solution structure to overcome the difficulty that RRAP has with both integer and real variables; two penalty functions for two objective functions to deal with solutions that violate constraints; a novel number-based self-adaptive update mechanism with a repository to store local nondominated solutions and replace the *gBest*; constrained nondominated-solution selection to prevent an excessive amount of disqualified and local nondominated solutions in the early generations; the crowding distance to force solutions cover the Pareto front uniformly. Moreover, the full-factorial design is implemented to determine the best combinations among eight different versions of MOSSO with different replacement policies, update strategies, and roles of *pBest*.

For four RRAPs adopted from the literature, the MOSSO outperformed NSGA-II and MOPSO, which are two well-known state-of-the-art methods, with regard to the Pareto plots and SP values. The weaknesses of these two state-of-art methods for solving such two-type variable bi-objective active

RRAP were discussed. The results provide important evidence that the SP and GD are insufficient for evaluating the performance of multi-objective Pareto algorithms. Thus, the proposed MOSSO can have better convergence and diversity of local nondominated solutions than that of NSGA-II and MOPSO, according to the analysis and comparisons. Our results encourage the extension of MOSSO to larger-scale and practical multi-objective problems.

**Acknowledgements**

This research was supported in part by the Ministry of Science and Technology, R.O.C. under grant MOST 102-2221-E-007-086-MY3 and MOST 104-2221-E-007-061-MY3.